\documentclass[10pt, a4paper]{article}

\usepackage{lrec-coling2024} 
\usepackage{inconsolata}
\usepackage{amsmath, amssymb}
\usepackage{booktabs}
\usepackage{graphicx}
\usepackage{multirow}
\usepackage{float, subfig}
\usepackage{xspace}
\usepackage[ruled, linesnumbered]{algorithm2e}

\newcommand{\modelname}{HiAdv\xspace}

\title{Utilizing Local Hierarchy with Adversarial Training for Hierarchical Text Classification}

\name{Zihan Wang, Peiyi Wang, Houfeng Wang$^{\ast}$} 

\address{National Key Laboratory of Multimedia Information Processing  \\
         School of Computer Science, Peking University \\
         \{wangzh9969, wangpeiyi9979\}@gmail.com,  wanghf@pku.edu.cn\\
}

\abstract{
Hierarchical text classification (HTC) is a challenging subtask of multi-label classification due to its complex taxonomic structure. Nearly all recent HTC works focus on how the labels are structured but ignore the sub-structure of ground-truth labels according to each input text which contains fruitful label co-occurrence information. In this work, we introduce this local hierarchy with an adversarial framework. We propose a \modelname framework that can fit in nearly all HTC models and optimize them with the local hierarchy as auxiliary information. We test on two typical HTC models and find that \modelname is effective in all scenarios and is adept at dealing with complex taxonomic hierarchies. Further experiments demonstrate that the promotion of our framework indeed comes from the local hierarchy and the local hierarchy is beneficial for rare classes which have insufficient training data.
 \\ \newline  }

\begin{document}

\maketitleabstract

\renewcommand{\thefootnote}{\fnsymbol{footnote}}

\footnotetext[1]{Corresponding author.}
\renewcommand{\thefootnote}{\arabic{footnote}}

\section{Introduction}
Hierarchical text classification (HTC) aims to categorize a text sample into a set of labels that are organized as a structured hierarchy \cite{silla2011survey}. Various multi-class classification problems can be extended to HTC by giving a pre-defined label hierarchy, such as scientific documents classification \cite{lewis2004rcv1, sadat-caragea-2022-hierarchical} or news categorization \cite{kowsari2017hdltex}. As the main difference to ordinary multi-class classification problems, how to utilize the large-scale, imbalanced label hierarchy is the key challenge of HTC \cite{mao2019hierarchical}.

Nearly all of the recent HTC works modeled the label hierarchy with a graph encoder: they feed in either both text and label information for a mixture representation \cite{zhou2020hierarchy, wang-etal-2022-hpt} or the hierarchy solely for a graph representation which is then fused with text representation  \cite{zhou2020hierarchy, deng2021htcinfomax, zhao2021hierarchical}. However, these works mainly focus on the constant global hierarchy but ignore the subgraph corresponding to each input text, which can contain structured label co-occurrence information \cite{jiang-etal-2022-exploiting}. This so-called local hierarchy is first introduced by \citet{wang2022incorporating} to generate positive samples for contrastive learning and they observe small improvements. However, this work considers the local hierarchy as flat but ignores its structure. Following this work, \citet{jiang-etal-2022-exploiting} takes further advantage of local hierarchy by a sequence-to-sequence approach. In their method, the local hierarchy is utilized through teacher forcing, which takes a ground-truth label sequence as input and predicts the next step of that sequence.
However, the transition from a graph to a sequence still loses some hierarchical information. Besides, their method generates labels layer by layer, so this method requires multiple inference times compared to other works.

\begin{figure}[t]
    \centering
    \includegraphics[width=0.9\linewidth]{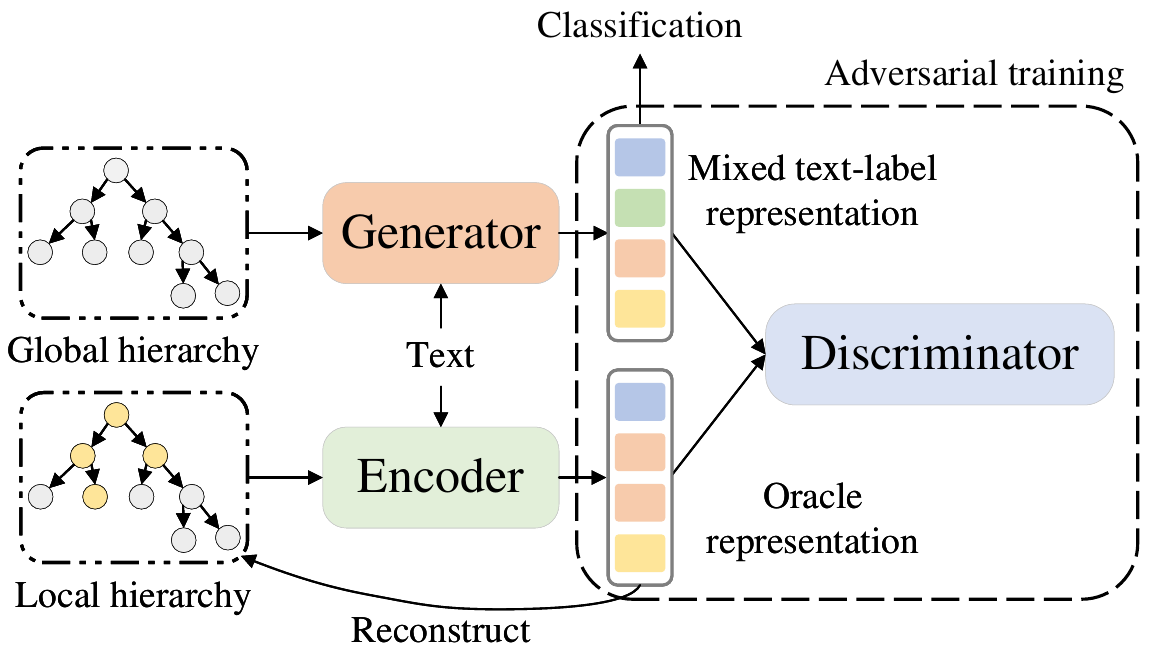}
    \caption{A demonstration of our adversarial framework. A generator and an encoder use global and local hierarchy as input respectively, and the output representations are trained adversarially.}
    \label{fig:intro}
\end{figure}

Is there a way that can directly model the local hierarchy as a graph? Stepping forward from the contrastive learning of \citet{wang2022incorporating}, a simple idea will supply: we can train the original text representation to be similar to a local-hierarchy-incorporated representation. In contrast to contrastive learning which requires dissimilar example pairs and previous work ignores totally, in this paper, we propose a hierarchy-aware adversarial framework (\modelname) to incorporate local hierarchy.

In detail, during training, our \modelname learns two representations adversarially. As shown in Figure \ref{fig:intro}, a generator inputs the text and global hierarchy and outputs a mixed text-label representation. Besides, an encoder generates an oracle representation by giving the ground-truth labels as input. In order to obtain high-quality local-hierarchy-related representation, the encoder organizes as an autoencoder-like structure: it encodes the local hierarchy to a representation and then reconstructs the representation back to the local hierarchy. A discriminator is then asked to distinguish the raw representation from the one with the local hierarchy. By encouraging the generator to fool the discriminator, the raw representation should be similar enough to the one with the local hierarchy after training. Furthermore, this idea does not rely on specific model architecture: we can apply it to any HTC model that involves a graph encoder.

We summarize our contributions as follows:
\begin{itemize}
    \item We propose an adversarial framework, \modelname, for HTC to incorporate local hierarchy. This is the first attempt to incorporate the entire local hierarchy.
    \item The framework does not rely on specific architecture so it can adapt to most existing HTC models which rely on a graph encoder to encode label hierarchy.
    \item Experiments demonstrate that our framework can constantly improve the performance of basic models on three datasets and achieve new state-of-the-art with the latest architecture as a backbone. We release our code at \href{https://github.com/wzh9969/HiAdv}{https://github.com/wzh9969/HiAdv}.
\end{itemize}

\begin{figure*}[ht]
    \centering
    \subfloat[HiBERT]{
    \label{fig:basic:a}
        \includegraphics[width=0.4\linewidth]{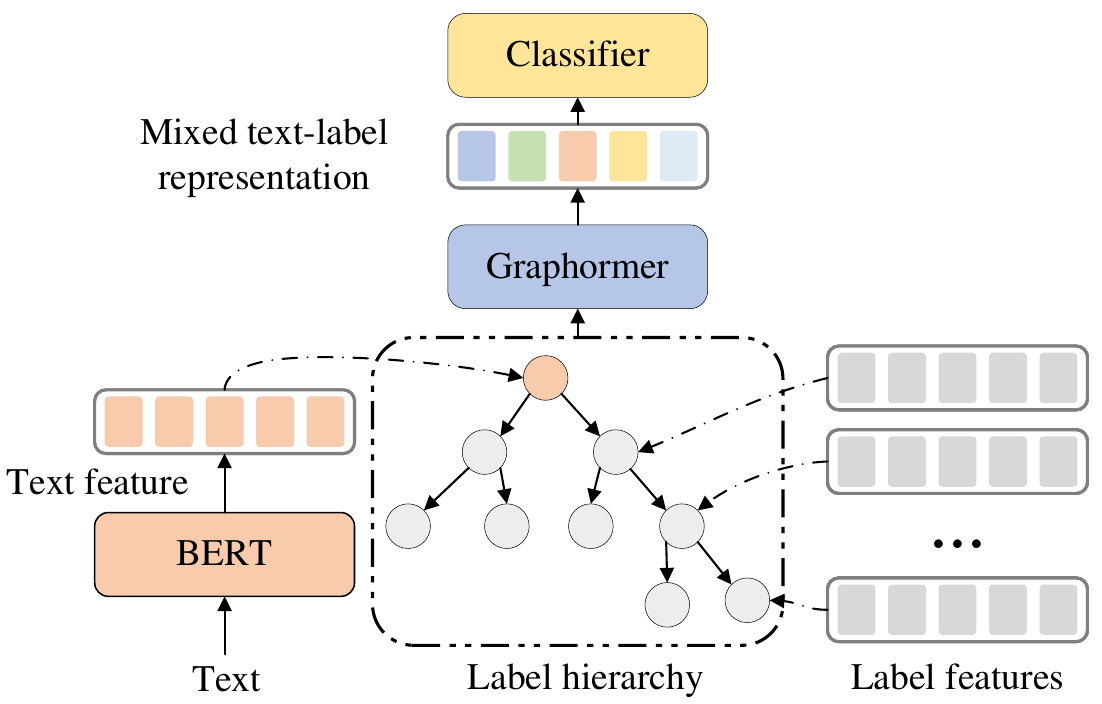}
    }
    \subfloat[HPT]{
    \label{fig:basic:b}
        \includegraphics[width=0.25\linewidth]{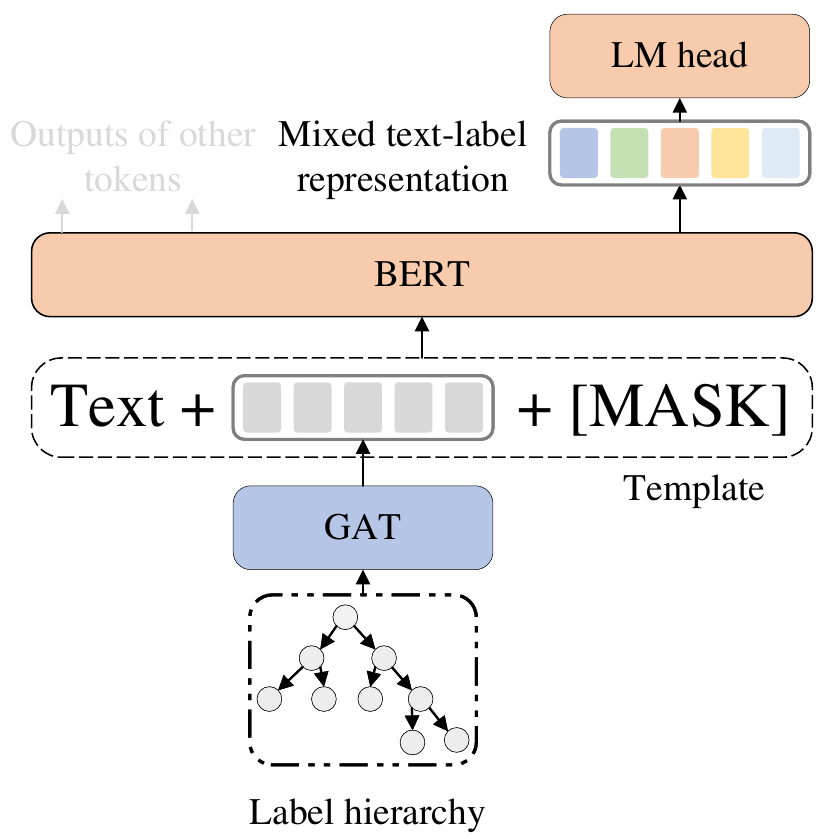}
    }
        \subfloat[An abstract model]{
    \label{fig:basic:c}
        \includegraphics[width=0.28\linewidth]{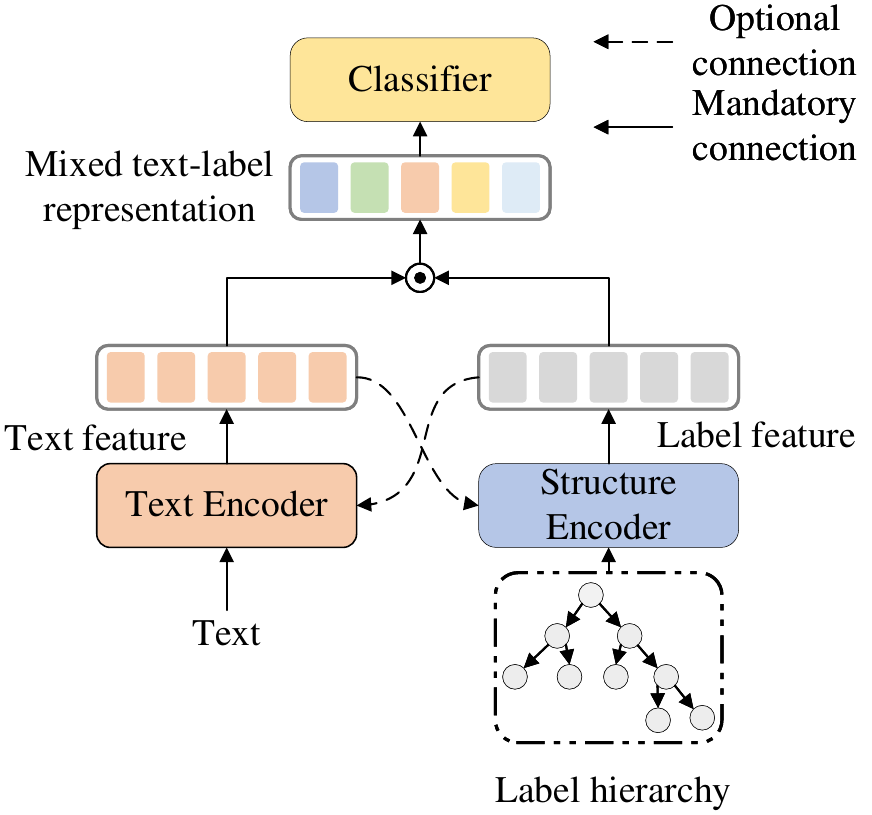}
    }
    \caption{
    Two HTC models and an abstract HTC model. (a) HiBERT. Feed BERT output text representation into a graph encoder. (b) HPT. A prompt tuning model with a hierarchy-aware template. (c) An abstract model. Two encoders dealing with text and structure respectively and a mixture mechanism for a mixed representation.
    }
    \label{fig:basic}
\end{figure*}

\section{Related Work}
\subsection{Hierarchy text classification}
Hierarchical text classification (HTC) is a challenging task due to its large-scale, imbalanced, and structured label hierarchy \cite{mao2019hierarchical}.
Recent works for HTC focus on global approaches, which build only one classifier for the entire graph \cite{zhou2020hierarchy}. 
The early global approaches neglect the hierarchical structure of labels and view the problem as a flat multi-label classification \cite{johnson2014effective}. After various attempts to coalesce the label structure \cite{wu2019learning, mao2019hierarchical, zhang2020hcn}, \citet{zhou2020hierarchy} demonstrate that encoding the holistic label structure directly by a structure encoder can most improve performance. Following this research, a bunch of models try to study how the structure encoder should interact with the text \cite{chen2020hyperbolic, chenhierarchy, deng2021htcinfomax, zhao2021hierarchical, zhu-etal-2023-hitin}. Besides the direct text-label interaction, \citet{wang-etal-2022-hpt} adopts prompt tuning for HTC to better utilize pretrained language models, and they predict labels layer-wisely to manually incorporate the depth information of the label hierarchy.
Despite a different training strategy, this work still follows the principle of integrating the output of the structure encoder into the text representation. \citet{ji-etal-2023-hierarchical} studies the problem under few-shot settings.

While previous works focus on the global hierarchy (i.e., the predefined label hierarchy), a few recent works claim the importance of local hierarchy (i.e., the hierarchy corresponding to each input text).
\citet{wang2022incorporating} adopts contrastive learning to directly inject hierarchical knowledge into the text encoder. In this work, ground-truth labels serve as the guidance to select classification-related tokens, which are then constructed as positive samples for contrastive learning. \citet{jiang-etal-2022-exploiting} views the problem as sequence generation: labels at different hierarchy layers are generated step by step. From this perspective, the local hierarchy is introduced naturally: only the labels in the local hierarchy should be generated. However, this method gets rid of the structure encoder entirely, which has been proven to be effective in all of the previous works.

\subsection{Adversarial Training}
Adversarial training is originally proposed in the Generative Adversarial Network (GAN) for image generation \cite{goodfellow2020generative}. It adversarially trains
a discriminator against a generator: the discriminator aims to distinguish real images from generated ones while the generator struggles to fool the discriminator. In NLP, adversarial networks are also applied in some generation tasks. For sequence generation, adversarial training serves as an alternative to step-by-step generation to avoid exposure bias issues \cite{scialom2020coldgans, liu2020catgan, chai2023improved}. For NLU tasks, recent studies show that adversarial training can be well-deployed on pre-trained language models \cite{zhufreelb, jiang2020smart, wanginfobert, wu2023toward}. Some works of neural topic modeling introduce adversarial networks to learn topic distributions \cite{wang2019atm, wang2020neural, hu2020neural}. Both \citet{wu2021conditional} and \citet{wu2022co} adopt adversarial training for multi-domain text classification to extract shared features across domains. However, the usage of adversarial training in their works focuses on multi-domain data and has little relation to the classification task itself. As a result, these methods cannot be adopted for general text classification or the hierarchical text classification that we are studying.

\section{Preliminary}
\subsection{Problem Definition}
For each hierarchical text classification (HTC) dataset, we have a predefined label hierarchy (i.e., global hierarchy) $\mathcal{H}=(\mathcal{Y}, E)$, where $\mathcal{Y}$ is the label set (also the node set of $\mathcal{H}$) and $E$ is the edge set. $\mathcal{H}$ is organized as a Directed Acyclic Graph (DAG) but we focus on a setting where every node except the root has one and only one father so that the hierarchy can be simplified as a tree-like structure.
Given an input text \textbf{x}, the models aim to categorize it into a label set (i.e., local hierarchy) $Y \subseteq \mathcal{Y}$.
The predicted label set $Y$ corresponds to one or more paths in $\mathcal{H}$. Each path starts from the root node and ends at any node on the tree.

\subsection{Basic Models} \label{sec:models}
Our framework is model-unrelated, so it is necessary to introduce two basic architectures we adopt in this work.

\subsubsection{HiBERT}
A text encoder with an auxiliary structure encoder is one of the simplest HTC models. Following previous work \cite{wang2022incorporating}, we adopt BERT \cite{devlin2018bert} as the text encoder and Graphormer \cite{ying2021transformers} as the structure encoder. We name this architecture Hierarchy-aware BERT (HiBERT) for simplicity.

Specifically, as in Figure \ref{fig:basic:a}, given a text input \textbf{x}, BERT encodes it to a text representation,
\begin{equation} \label{eq:1}
    \mathbf{h}_{\mathrm{text}}=\mathrm{BERT}(\textbf{x})
\end{equation}
The structure encoder then takes the label hierarchy and text representation as input and outputs a mixed representation.
To utilize label information, we use the average of BERT token embedding of the label name as its label embedding $\mathbf{l}_i$. For the root node which has no particular meaning, we use the text representation as label embedding. We use the output graph representation of the root node as a mixed representation of text and labels:
\begin{equation}
    \mathbf{h}_{\mathrm{mix}}=\mathrm{Graphormer}([\mathbf{h_{text}},\mathbf{L}], \mathcal{H})_{\mathrm{root}}
\end{equation}
Finally, a linear classifier is used to calculate the probability of each class:
\begin{equation}
    P=\mathrm{sigmoid}(\mathbf{W}_{c}\mathbf{h}_{\mathrm{mix}}+\mathbf{b}_c)
\end{equation}
where $\mathbf{W}_{c}$ and $\mathbf{b}_c$ are weight matrix and bias.

\subsubsection{Hierarchy Prompt Tuning}
Hierarchy Prompt Tuning (HPT) \cite{wang-etal-2022-hpt} is the state-of-the-art model which regards the problem as a masked language model problem. As shown in Figure \ref{fig:basic:b}, HPT first encodes label hierarchy into prompting features $\mathbf{T}$ by a GAT:
\begin{equation}
    \mathbf{T}=\mathrm{GAT}(\mathbf{L},\mathcal{H})
\end{equation}
where $\mathbf{L}=\{\mathbf{l}_i\}$ is the label embedding that contains node information.
Then, the input text \textbf{x} and prompting features $\mathbf{T}$ are filled into a template, where a $\mathrm{[MASK]}$ position is reserved as in the masked language model (MLM) task. BERT then encodes the template but we focus only on the output representation of the $\mathrm{[MASK]}$ token.
By calculating the probability of filling each label embedding into the $\mathrm{[MASK]}$ position, HPT predicts in an MLM manner. We define the BERT output of the $\mathrm{[MASK]}$ position as mixed text-label representation $\mathbf{h}_\mathrm{mix}$ in accordance with HiBERT. Please refer to the original paper for more details.

\section{Methodology}
In this section, we introduce \modelname. The whole framework is illustrated in Figure \ref{fig:adv}. Our framework can be applied to any HTC architecture that involves a text encoder and a graph encoder.

\subsection{Abstraction of HTC Models}
Despite differing in architecture details, both preceding models as well as most other architectures HTC works selected can be abstracted to a text encoder $E_\mathrm{text}$ and a structural encoder $E_\mathrm{structure}$:
\begin{equation} \label{eq:abs}
    \begin{aligned}
        \mathbf{h}_\mathrm{text} &= E_\mathrm{text}(\textbf{x}, \mathbf{h}_\mathrm{label}) \\
        \mathbf{h}_\mathrm{label} &= E_\mathrm{structure}(\mathbf{L}, \mathcal{H}, \mathbf{h}_\mathrm{text})
    \end{aligned}
\end{equation}
As in Equation \ref{eq:abs} and Figure \ref{fig:basic:c}, the text encoder takes text \textbf{x} as input while the structural encoder is built according to the label hierarchy $\mathcal{H}$ and takes label embedding $\mathbf{L}$ as input. Besides, one encoder may utilize the output representation of the other. Each encoder outputs a representation in $\mathit{R}^{d}$.

Finally, a mixture mechanism coalesces two representations:
\begin{equation}\label{eq:hmix}
    \mathbf{h}_\mathrm{mix} = \mathbf{h}_\mathrm{text} \odot \mathbf{h}_\mathrm{label}
\end{equation}
and a multi-label classifier $C$ predicts probability distributions:
\begin{equation}\label{eq:ref}
    P = \mathrm{sigmoid}(C(\mathbf{h}_\mathrm{mix}))
\end{equation}

During training, a multi-label classification loss such as binary cross entropy is adopted:
\begin{equation}\label{eq:lc}
    L_\mathrm{C}=Loss(P, Y)
\end{equation}

This abstract model generates a representation $\mathbf{h}_\mathrm{mix}$ according to a given input $\mathbf{x}$, so we can define it as a generator under the perspective of adversarial training.

\subsection{Encoder Network}
In the context of adversarial training, an encoder network encodes a real input $\hat{\mathbf{x}}$ into a real distribution. In our scenario, the encoder network attempts to generate an oracle representation that takes local hierarchy as input and can perform the best classification. We organize the encoder as an autoencoder: it encodes local hierarchy along with text into the oracle representation and then reconstructs the local hierarchy.

\begin{figure}[t]
    \centering
    \includegraphics[width=0.93\linewidth]{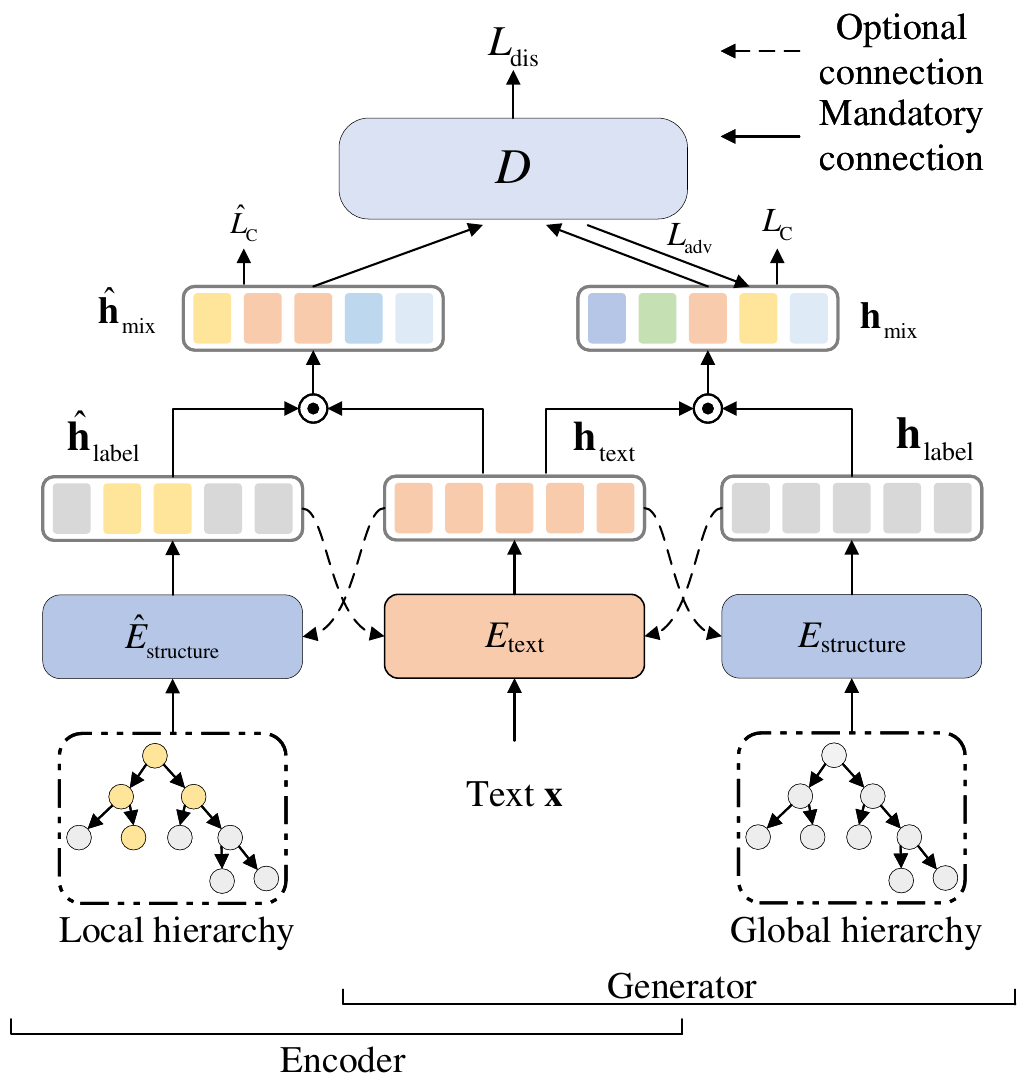}
    \caption{A demonstration of our adversarial framework. The generator and the encoder share the same text encoder. We omit the classifier for clarity, which takes $\mathbf{h}_\mathrm{mix}$ and $\hat{\mathbf{h}}_\mathrm{mix}$ for input to classify and generates classification losses $L_\mathrm{C}$ and $\hat{L}_\mathrm{C}$ during training.}
    \label{fig:adv}
\end{figure}

To incorporate local hierarchy, we modify the label embedding in Equation \ref{eq:abs}. For label $y_i$, we add an oracle label embedding to indicate whether this label is in the local hierarchy $Y$:
\begin{equation}
    \begin{aligned}
        \hat{\mathbf{l}}_i=\mathbf{l}_i+\left\{
        \begin{aligned}
         & \mathbf{e}_1, &  y_i \in Y \\
         & \mathbf{e}_0, &  y_i \not\in Y \\
        \end{aligned}
        \right. 
    \end{aligned}
\end{equation}
where $\mathbf{e}_0$ and $\mathbf{e}_1$ are two learnable embeddings that indicate whether $y_i$ is in $Y$.
As a result, the label embedding matrix $\hat{\mathbf{L}}=\{\hat{\mathbf{l}}_i\}$ contains the local hierarchy as a prior. As in Figure \ref{fig:adv}, we use an independent structure encoder to deal with the local hierarchy while other components of the encoder share with the generator:
\begin{equation}\label{eq:hmixhat}
\begin{aligned}
        \hat{\mathbf{h}}_\mathrm{label} &= \hat{E}_\mathrm{structure}(\hat{\mathbf{L}}, \mathcal{H}, \mathbf{h}_\mathrm{text}) \\
        \hat{\mathbf{h}}_\mathrm{mix} &= \mathbf{h}_\mathrm{text} \odot \hat{\mathbf{h}}_\mathrm{label}
\end{aligned}
\end{equation}
where $\hat{\mathbf{h}}_\mathrm{mix}$ is the demanded oracle representation which takes the local hierarchy as a prior.

The oracle representation $\hat{\mathbf{h}}_\mathrm{mix}$ is then reconstructed to the local hierarchy $Y$ with the classifier $C$, which is similar to an autoencoder that aims to find the best representation for reconstructing the local hierarchy.
During training, $\hat{\mathbf{h}}_\mathrm{mix}$ is guided by the classification loss the same way as $\mathbf{h}_\mathrm{mix}$ in the generator:
\begin{equation}\label{eq:lchat}
    \hat{L}_\mathrm{C}=Loss(\mathrm{sigmoid}(C(\hat{\mathbf{h}}_\mathrm{mix})), Y)
\end{equation}

\begin{algorithm}[t]
	\caption{Framework of \modelname.}
	\label{alg:1}
	\KwIn{Training dataset $T_\mathrm{train}=\{(\textbf{x}, Y)\}$, global hierarchy $\mathcal{H}$, test dataset $T_\mathrm{test}=\{\textbf{x}\}$}
	\KwOut{Model prediction $\{\hat{Y}\}$}
 \tcp{Initialization}
 
	Initialize $E_\mathrm{text}$, $E_\mathrm{structure}$, $\hat{E}_\mathrm{structure}$, $C$, and $D$ randomly\;
	 \tcp{Training}
	\While{\textnormal{not converged}}{
        Sample a data point $(\textbf{x}, Y)$ from $T_\mathrm{train}$\;
 
		Calculate $\mathbf{h}_\mathrm{mix}$ and $\hat{\mathbf{h}}_\mathrm{mix}$ by Equation \ref{eq:hmix} and \ref{eq:hmixhat}\;
    \tcp{Update $D$}
  Fill $\mathbf{h}_\mathrm{mix}$ and $\hat{\mathbf{h}}_\mathrm{mix}$ successively into Equation \ref{eq:disp} to obtain $p$ and $\hat{p}$\;
  Calculate $L_\mathrm{dis}$ and $\hat{L}_\mathrm{dis}$ by Equation \ref{eq:dis}\;
  Back-propagate $L_\mathrm{dis}+\hat{L}_\mathrm{dis}$ and update parameters of the discriminator\;
    \tcp{Update the rest}
  Calculate classification loss $L_\mathrm{C}$ and $\hat{L}_\mathrm{C}$ by Equation \ref{eq:lc} and \ref{eq:lchat}\;
  Calculate adversarial loss $L_\mathrm{adv}$ by Equation \ref{eq:adv}\;
  Back-propagate $L_\mathrm{C}+\hat{L}_\mathrm{C}+L_\mathrm{adv}$ and update parameters of all components except the discriminator\;
	}
	\tcp{Testing}
	 \ForEach{\textnormal{Text} $\mathbf{x}\in T_\mathrm{test}$}{
  Calculate $P$ by Equation \ref{eq:ref} using $E_\mathrm{text}$, $E_\mathrm{structure}$, and $C$;
  
  Select $\hat{Y}=\{P_{i}>\tau\ \forall i\}$;
     }

\end{algorithm}

\subsection{Discriminator}
With a generated representation $\mathbf{h}_\mathrm{mix}$ and an oracle representation $\hat{\mathbf{h}}_\mathrm{mix}$, a discriminator $D$ attempts to distinguish whether a representation $\mathbf{h}$ is generated with or without the local hierarchy. This is a binary classification problem so we use a simple two-layer linear classifier as the discriminator:
\begin{equation}\label{eq:disp}
    p=\mathrm{Sigmoid}(\mathbf{W}_2\mathrm{ReLU}(\mathbf{W}_1\mathbf{h}+\mathbf{b}_1)+b_2)
\end{equation}
where $\mathbf{W}_1\in\mathit{R}^{d\times d}$, $\mathbf{W}_2\in\mathit{R}^{1\times d}$, $\mathbf{b}_1\in\mathit{R}^d$, and $b_2\in\mathit{R}$ are weights and bias.

The discriminator is guided by binary cross-entropy loss:
\begin{equation}\label{eq:dis}
    L_{dis}=-(I\log p+(1-I)\log (1-p))
\end{equation}
where $I$ is an indicator
\begin{equation}
    \begin{aligned}
        I=\left\{
        \begin{aligned}
         & 0, &  \mathbf{h}=\mathbf{h}_\mathrm{mix} \\
         & 1, &  \mathbf{h}=\hat{\mathbf{h}}_\mathrm{mix} \\
        \end{aligned}
        \right. 
    \end{aligned}
\end{equation}

\subsection{Adversarial Framework}
The generator, the encoder, and the discriminator train adversarially. As mentioned, the discriminator aims to distinguish the representation while the generator attempts to fool the discriminator. As shown in Figure \ref{fig:adv}, besides the classification loss, the generator is guided by an adversarial loss:
\begin{equation} \label{eq:adv}
    L_{adv} = -(1-I)\log p
\end{equation}
Notice that $L_{adv}$ only updates the parameters of the generator but does not affect the parameters of the encoder or that of the discriminator so that we only consider $\mathbf{h}=\mathbf{h}_\mathrm{mix}$ in Equation \ref{eq:adv}.

During training, both the generator and the encoder take the same text \textbf{x} as input and outputs $\mathbf{h}_\mathrm{mix}$ and $\hat{\mathbf{h}}_\mathrm{mix}$ respectively. The discriminator then calculates $p$ so that every loss can be computed. After training, the framework degrades to a single generator, which can predict independently as Equation \ref{eq:ref}. Since the problem is a multi-label classification, we select labels that have probabilities greater than a threshold $\tau$ as model predictions. The whole process is illustrated in Algorithm \ref{alg:1}.

\section{Experiments}
\subsection{Experiment Setup}

\paragraph{Datasets and Evaluation Metrics}
We experiment on Web-of-Science (WOS) \citeplanguageresource{kowsari2017hdltex}, NYTimes (NYT) \citeplanguageresource{sandhaus2008new}, and RCV1-V2 \citeplanguageresource{lewis2004rcv1} datasets for analysis. The statistic details are illustrated in Table \ref{tab:1}.
We follow the data processing of previous work \cite{zhou2020hierarchy} and measure the experimental results with Macro-F1 and Micro-F1.

\begin{table}[t]
\resizebox{\linewidth}{!}{
\begin{tabular}{ccccccc}
\toprule
Dataset & $|Y|$   & Depth &   Avg($|y_i|$)   & Train  & Dev   & Test    \\ \midrule
WOS     & 141 & 2     & 2.0  & 30,070 & 7,518 & 9,397   \\
NYT     & 166 & 8     & 7.6  & 23,345 & 5,834 & 7,292   \\
RCV1-V2    & 103 & 4     & 3.24 & 20,833 & 2,316 & 781,265 \\ \bottomrule
\end{tabular}
}
\caption{Data statistics. $|Y|$ is the number of classes. Depth is the maximum level of hierarchy. Avg($|y_i|$) is the average number of classes per sample.}
\label{tab:1}
\end{table}

\begin{table*}[t]
\centering
\resizebox{\linewidth}{!}{
\begin{tabular}{lcccccc}
\toprule
\multirow{2}{*}{Model}                                & \multicolumn{2}{c}{WOS (Depth 2)} & \multicolumn{2}{c}{RCV1-V2 (Depth 4)} & \multicolumn{2}{c}{NYT (Depth 8)} \\ \cmidrule(l){2-7} 
                                                      & Micro-F1        & Macro-F1        & Micro-F1          & Macro-F1          & Micro-F1        & Macro-F1       \\  \midrule
BERT \cite{wang2022incorporating}            & 85.63      & 79.07      & 85.65        & 67.02      & 78.24      & 66.08        \\
BERT+HiAGM\cite{wang2022incorporating}                                 & 86.04           & 80.19           & 85.58             & 67.93             & 78.64           & 66.76          \\
BERT+HTCInfoMax\cite{wang2022incorporating}                            & 86.30           & 79.97           & 85.53             & 67.09             & 78.75               & 67.31              \\
BERT+HiMatch \cite{chenhierarchy}    & 86.70           & 81.06           & 86.33             & 68.66             & -               & -              \\
 HGCLR \cite{wang2022incorporating}   & 87.11           & 81.20           & 86.49           & 68.31            & 78.86              & 67.96              \\
HPT \cite{wang-etal-2022-hpt}          & 87.16           & 81.93           & 87.26             & 69.53         
                                                        & 80.42           & 70.42          \\ 
 HBGL \cite{jiang-etal-2022-exploiting} & \textbf{87.36} & \textbf{82.00} & 87.23 & \textbf{71.07} & 80.47 & 70.19 \\
 \midrule
HiBERT & 85.77 & 80.10 & 86.49 & 68.82 & 79.49 & 68.40 \\
HiBERT + \modelname$^\dagger$ & 86.38 & 80.78 & 86.74 & 69.43&79.56 & 69.30\\
 HPT* & 87.08 & 81.59 & 86.96 & 69.25  &80.21&70.14 \\
 HPT* + \modelname$^\dagger$ & 87.20 & 81.62 & \textbf{87.36} & 69.62& \textbf{80.83} & \textbf{70.78}\\
 \bottomrule
\end{tabular}
}
\caption{F1 scores on $3$ datasets. The best results are in boldface. *In order to fit in \modelname, we modify the predicting strategy of HPT so the result is slightly lower. Hereinafter we still denote it as HPT for simplicity. $^\dagger$Improvements are statistical significant with $p<0.05$.
}

\label{tab:2}
\end{table*}

\paragraph{Baselines}
For systematic comparisons, we introduce a variety of HTC baselines.

\begin{itemize}
    \item \textbf{BERT} \cite{devlin2018bert}. A widely used pretrained language model that can serve as a text encoder. All of the other baselines as well as our method are based on BERT.
    \item \textbf{HiAGM} \cite{zhou2020hierarchy}, \textbf{HTCInfoMax} \cite{deng2021htcinfomax}, and \textbf{HiMatch} \cite{chenhierarchy}. These three methods all encode text and taxonomic hierarchy separately and propose different migration strategies.
    \item \textbf{HGCLR} \cite{wang2022incorporating}. HGCLR regulates BERT representation by contrastive learning. It only introduces nodes of the local hierarchy but ignores how they are connected.
    \item \textbf{HPT} \cite{wang-etal-2022-hpt}. HPT proposes a new architecture with prompt tuning. The text encoder of HPT encodes text and the output of the graph encoder simultaneously.
    \item \textbf{HBGL} \cite{jiang-etal-2022-exploiting}. HBGL introduces local hierarchy in a sequence-to-sequence manner, differing from other works which are discriminative models. The architecture of HBGL is a pure BERT so we cannot adopt it as a backbone.
\end{itemize}

\paragraph{Implement Details} 
We implement our model using PyTorch. Following previous work \cite{chenhierarchy}, we use \texttt{bert-base-uncased} as our text encoder. We test on two architectures introduced in Section \ref{sec:models}. For HPT, since the adversarial framework can only focus on one representation, we modify the predicting strategy so that the model predicts with one single representation instead of predicting layer-wisely. For HiBERT, we adopt Zero-bounded Log-sum-exp \& Pairwise Rank-based \cite{su2022zlpr} that HPT uses, which has proven to be a more suitable loss function than binary cross-entropy for HTC. We use a batch size of $8$ to fill one Nvidia RTX 3090 (24G) fully. All other hyper-parameters follow HPT \cite{wang-etal-2022-hpt}. For more stable training, we train the model without the adversarial loss for the first epoch and stop training if Macro-F1 on the development set does not increase for $5$ epochs. Following \citet{wang-etal-2022-hpt}, we select the checkpoint with maximum Macro-F1 on the development set as the final model and report the best results among $5$ individual runs.

\begin{table*}[ht]
\centering
\resizebox{\textwidth}{!}{%
\begin{tabular}{lcccccc}
\toprule
\multirow{2}{*}{Variants}& \multicolumn{2}{c}{WOS (Depth 2)} & \multicolumn{2}{c}{RCV1-V2 (Depth 4)} & \multicolumn{2}{c}{NYT (Depth 8)} \\ \cmidrule(l){2-7} 
                          & Micro-F1   & Macro-F1   & Micro-F1     & Macro-F1     & Micro-F1   & Macro-F1   \\ \midrule
HiBERT & 86.90 & 80.82 & 87.50 & 68.69 & 79.93 & 69.84 \\
HiBERT + \modelname   & \textbf{87.10}       & \textbf{81.36}       & \textbf{87.83}         & \textbf{69.66}         &\textbf{80.00}      & \textbf{70.68}      \\
- w/o adversarial loss              & 86.57       & 80.61       &  87.46    &   68.42  & 79.33      & 69.23      \\
HiBERT + Contrastive learning                  & 86.49       & 79.46       & 87.09         & 68.40         & 77.46      & 65.18      \\ \midrule
HPT & 87.68& 81.72 & 88.04 & 69.41 & 80.49 & 71.18\\
HPT + \modelname     & \textbf{87.84}       & \textbf{82.07}       & \textbf{88.40}         &\textbf{69.80}         & \textbf{80.99}      & \textbf{71.73}      \\
- w/o adversarial loss  & 87.40    &  81.61   &   88.03    &   69.34    & 80.66      & 71.16      \\
HPT + Contrastive Learning                  & 87.74       & 81.91       & 88.13         & 69.28        & 80.84       & 71.69       \\ \bottomrule
\end{tabular}%
}
\caption{Results of different optimizing strategies on the development set.  The best results are in boldface. We test two backbones with 1) no treatment; 2) the proposed \modelname; 3) the proposed \modelname but without the adversarial loss; 4) contrastive learning.}
\label{tab:adv}
\end{table*}

\subsection{Main Results}
Table \ref{tab:2} illustrates our main results. Among baseline methods, HBGL and HPT perform similarly on all datasets except the Macro-F1 of RCV1-V2, which we believe is the advantage of sequence-to-sequence generation.

The proposed \modelname steadily improves the performance of both backbone models on all datasets. WOS has a label hierarchy of two layers and the local hierarchy has only one path, making it the easiest dataset among all tested datasets. As a result, applying \modelname on HPT has little improvement because it has already learned the hierarchy information well without further treatment. HiBERT is a weaker model than HPT, so it can still benefit from \modelname on a simple dataset. As for NYT which has a hierarchy of $8$ layers and has the most intricate label structure, existing methods may not fully learn that information so that our adversarial framework achieves new state-of-the-art.

As mentioned, we use a variant of HPT by removing the layer-wise prediction. Although information about the depth of hierarchy is lost to some extent, the extra local hierarchy provided by the adversarial framework outweighs the loss except on WOS. WOS is a special case where the hierarchy is too easy to learn, so a handmade rule to force the model to predict layer-wise is sufficient.

\subsubsection{Discussion on Model Capacity}
Model capacity can affect the performance of \modelname. HPT is a stronger backbone than HiBERT and it leads to divergent behaviors between these two models.

When applying \modelname on HiBERT, the improvements of Micro-F1 descend according to the difficulty of label hierarchy while the improvements of Macro-F1 for HiBERT are steady. This phenomenon indicates that the local hierarchy benefits rare classes with insufficient training samples more than sufficiently trained classes when the model capacity is limited. However, for a stronger basic model HPT, besides the WOS dataset which the model has learned well enough, applying \modelname can improve both Micro-F1 and Macro-F1, showing that our adversarial framework is adept at dealing with complex structures for powerful models.

\subsubsection{Comparison with Other Local Hierarchy Methods}
When comparing our method with HGCLR and HBGL which also involves local hierarchy, the results demonstrate that our adversarial framework can better utilize the local hierarchy. 

If we use HPT, which ignores local hierarchy completely, as a baseline, both HGCLR and HBGL reveal better performance on WOS which has the simplest hierarchy but behaves relatively poorly on NYT which has the most complex hierarchy. These two methods view the local hierarchy as either flat or a sequence so that they lose more information about structure when the hierarchy is more complex. Besides, HGBL utilizes local hierarchy in a sequence-generation manner so there is a gap between the ground-truth local hierarchy input for training and model-predicted hierarchy input for testing. When the hierarchy is deep, this gap becomes more evident. On the contrary, the promotion of our \modelname has a positive correlation to the complexity of hierarchy: we observe the most improvement at NYT but nearly no improvement at WOS, showing that \modelname can better handle complex local hierarchy and learn more from it than previous methods.

\subsection{Analysis}

\subsubsection{Effect of Adversarial Training}
We select adversarial training to incorporate local hierarchy but it is not the only choice. \citet{wang2022incorporating} uses contrastive learning for a similar purpose so we compare \modelname with it here.
As in Table \ref{tab:adv}, we exhibit results on the development set of removing the adversarial loss and replacing \modelname with contrastive learning.

After removing the adversarial loss, \modelname degrades into a data augmentation method where the local-hierarchy-involved data generated by the encoder serve as augment data. Although random improvement can be observed, in most cases simple data augmentation is counteractive. Directly introducing local hierarchy will leak the ground-truth labels so that models tend to rely on the local hierarchy instead of learning from it without further guidance.

Contrastive learning produces promising results with HPT but fails to work with HiBERT. Contrastive learning directly optimizes the target representation and pulls it toward the oracle representation. Although robust models like HPT may gain from this optimization, weak models such as HiBERT experience countereffects. On the contrary, adversarial training is a more gentle approach that only requires two representations to be similar enough to fool a discriminator. As a result, \modelname is a more robust framework that can fit in more backbones.

\subsubsection{Effect of Local Hierarchy}
We further study how the local hierarchy affects \modelname on the NYT dataset due to its complex hierarchy. Table \ref{tab:ablation} enumerates the results of some variants on the development set of NYT. We study how the model reacts when providing a partial local hierarchy, no local hierarchy, and a wrong local hierarchy. We implement the partial local hierarchy by randomly dropping $15\%$ of labels and the wrong local hierarchy by randomly selecting the same amount of labels.

As is shown, our framework indeed learns from the local hierarchy. With it corrupted, results of both HiBERT and HPT drop significantly, showing that the promotion of \modelname indeed comes from the local hierarchy. As for removing the hierarchy partially or completely, variants with more local hierarchy have better performance. This observation is in accordance with the deduction from the main results that \modelname can learn more from complex hierarchy than simple one. With no local hierarchy or wrong hierarchy, the performance of HPT drops dramatically, which demonstrates that HPT is more sensitive to local hierarchy and thus more suitable for our framework from another perspective.

\begin{table}[t]
\centering
\resizebox{\linewidth}{!}{
\begin{tabular}{lcc}
\toprule
Variants                   & Micro-F1 & Macro-F1 \\ \midrule
HiBERT  + \modelname                    & \textbf{80.00}    &\textbf{70.68}    \\
- w/ partial local hierarchy & 79.51 & 69.64   \\
- w/ no local hierarchy & 78.46 & 68.09   \\
- w/ wrong local hierarchy  & 78.67 & 67.02     \\ \midrule
HPT + \modelname                  & \textbf{80.99}    & \textbf{71.73}    \\
- w/ partial local hierarchy & 80.86 & 71.50  \\
- w/ no local hierarchy  & 77.82 & 64.14   \\
- w/ wrong local hierarchy  & 77.09 & 61.87   \\ \bottomrule
\end{tabular}}
\caption{Results of modifying local hierarchy on the development set of NYT dataset. The best results are in boldface. We show the original results and the results when we input partial, empty, and wrong local hierarchy.
}
\label{tab:ablation}
\end{table}

Besides, changing the local hierarchy has a relatively lower impact on Micro-F1 than on Macro-F1, which we have observed similar behaviors in the main results. These results further demonstrate that the local hierarchy elevates the performance of rare classes more than normal classes.

\begin{figure}[t]
    \centering
    \subfloat[]{
    \label{fig:layers:a}
        \includegraphics[width=0.7\linewidth]{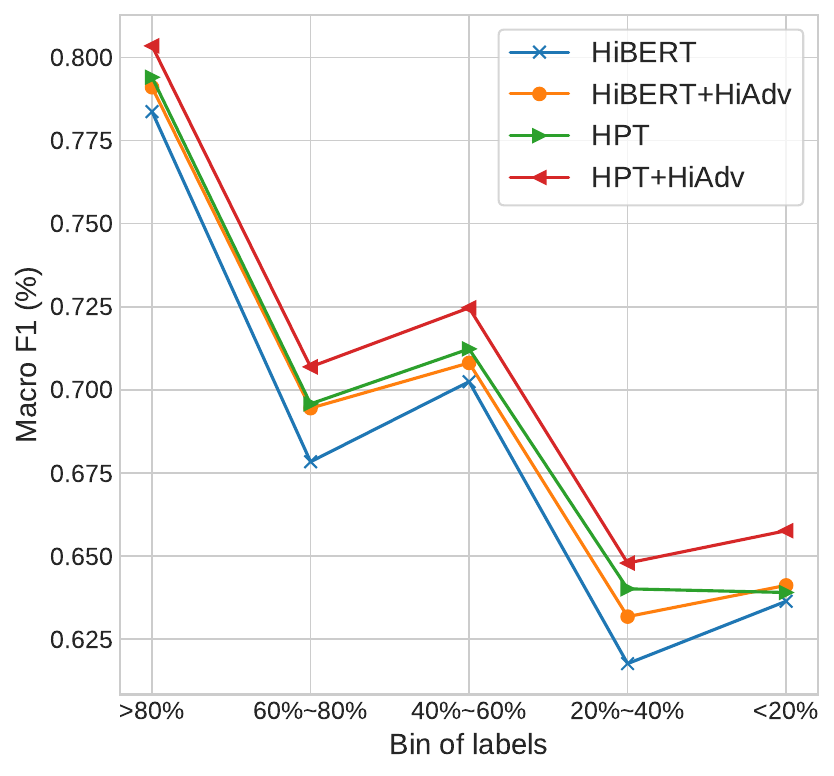}
    }
    \quad
    \subfloat[]{
    \label{fig:layers:b}
        \includegraphics[width=0.7\linewidth]{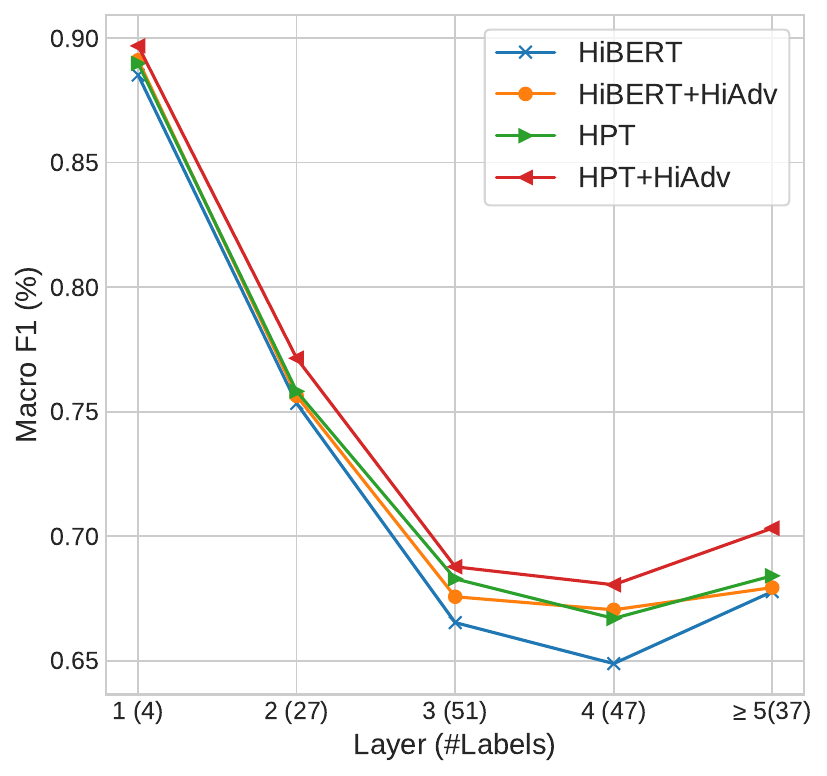}
    }
    \caption{Macro F1 scores of label clusters on the development set of NYT. (a) Label clusters grouped by the number of training samples. >80\% means this cluster of labels has training instances of more than 80\% of labels. The rest are arranged similarly. (b) Label clusters grouped by depth in the hierarchy. 
    }
    \label{fig:layers}
\end{figure}

\subsubsection{Results on Imbalanced Hierarchy}
One of the key challenges of hierarchical text classification is the imbalanced label hierarchy. As mentioned, our method affects Macro-F1 more than Micro-F1, showing that it can alleviate the imbalance issues to some extent. In this section, we visualize how our model resolves the issue of imbalance on the development set of NYT.

The imbalance can be viewed from two perspectives in HTC. For one, the number of training instances of each class is different so some classes may not be fully trained. As shown in Figure \ref{fig:layers:a}, we cluster labels into five bunches based on the number of training instances. \modelname is effective to all clusters and can promote rare classes such as "<20\%" cluster for HPT and "20\%-40\%" cluster for HiBERT.

For another, the number of labels at different hierarchy layers is different. As shown in Figure \ref{fig:layers:b}, \modelname mainly acts on deep layers (layers $\ge$ 4). The structure of label hierarchy is more complicated in the deep so this phenomenon demonstrates that our framework can utilize complex local hierarchy effectively. 

\subsubsection{Time Efficiency}
\modelname requires an additional graph encoder so it requires extra computational costs. Since our approach is a framework, the extra computational costs heavily depend on the backbone model. In this section, we study how \modelname affects the computational costs of different backbones.

In Table \ref{fig:time} we exhibit the rough training time consumed by one epoch on the NYT dataset. The straightforward implementation of \modelname requires roughly double the training time for both backbones. Compared to a single model that needs a one-time back-propagation, our framework requires two-time back-propagations for adversarial training. But from Figure \ref{fig:adv} it can be observed that both passes of back-propagation share the same BERT encoder, which contains most of the parameters so an optimization algorithm is easy to find.

For HiBERT where BERT only takes text as input (Equation \ref{eq:1}), the BERT encoder only needs one-time back-propagation exactly. As a result, the optimized \modelname requires around $7\%$ of extra training time. For HPT where BERT takes graph information as input, the second pass is necessary. But the input sequence of BERT is still mostly the same on both passes so it is possible to merge those two sequences into one. With that optimization implemented, \modelname only requires less than $20\%$ more training time.

During inference, our framework degrades into a single backbone model so it does not affect inference time. In practice, applying our framework needs no extra computational cost after deployment.

\begin{table}[t]
\begin{tabular}{lcc}
\toprule
Variants      & HiBERT & HPT  \\ \midrule
w/o \modelname                  & 560    & 690  \\
w/ \modelname, w/o optimization & 1020   & 1380 \\
w/ \modelname, w/ optimization  & 600    & 810  \\ \bottomrule
\end{tabular}
\caption{Rough seconds consumed on one epoch on the train set of NYT.
    }
    \label{fig:time}
\end{table}

\section{Conclusion}
In this paper, we propose a hierarchy-aware adversarial framework (\modelname) for assisting existing HTC models to incorporate local hierarchy. For any HTC model that involves a text encoder and a graph encoder, \modelname treats it as a generator while an extra encoder encodes the local hierarchy into an oracle representation. A discriminator tries to distinguish the original representation from the oracle representation while the generator attempts to fool the discriminator. Experiments show that the adversarial framework is adept at dealing with complex hierarchies or promoting weak models that cannot fully learn the hierarchy. Further experiments demonstrate that the effect of \modelname comes from the local hierarchy and the local hierarchy is beneficial for classes with deficient training instances.

\section{Acknowledgements}
We thank all the anonymous reviewers for their constructive feedback. This work was supported by the National Natural Science Foundation of China (62036001) and the National Science and Technology Major Project (2022ZD0116308).

\section{Bibliographical References}\label{sec:reference}

\bibliographystyle{lrec-coling2024-natbib}
\bibliography{lrec-coling2024-example}

\section{Language Resource References}
\label{lr:ref}
\bibliographystylelanguageresource{lrec-coling2024-natbib}
\bibliographylanguageresource{languageresource}

\end{document}